\documentclass[letterpaper, 10 pt, conference]{ieeeconf}  

\IEEEoverridecommandlockouts                              
\usepackage[utf8]{inputenc}
\usepackage[T1]{fontenc}
\usepackage{times}
\usepackage{epsfig}
\usepackage{graphicx}
\usepackage{amsmath}
\usepackage{amssymb}
\usepackage{graphics} 
\usepackage{mathtools}
\usepackage{subcaption}
\usepackage{booktabs}
\usepackage{adjustbox}
\usepackage{caption}
\usepackage{stfloats}
\usepackage{multirow}
\usepackage[linesnumbered,ruled,lined,commentsnumbered]{algorithm2e}
\usepackage[colorlinks]{hyperref}
\usepackage{algpseudocode}
\usepackage{rotating}
\usepackage{tabularray}
\usepackage{rotating}
\usepackage{makecell}
\usepackage{multirow}
\newcommand\inv[1]{#1\raisebox{1.15ex}{$\scriptscriptstyle-\!1$}}
\usepackage{makecell}
\usepackage{array}
\usepackage{multirow}

\title{\LARGE \bf RecNet: An Invertible Point Cloud Encoding through Range Image Embeddings for Multi-Robot Map Sharing and Reconstruction}

\author{Nikolaos Stathoulopoulos, Mario A.V. Saucedo, Anton Koval and George Nikolakopoulos 
\thanks{The Authors are with the Robotics and AI Group, Department of Computer, Electrical and Space Engineering, Lule\r{a} University of Technology, 971 87 Lule\r{a}, Sweden. Corresponding Author's Email: \texttt{niksta@ltu.se}}
\thanks{This work has been partially funded by the European Union's Horizon 2020 Research and Innovation Programme, under the Grant Agreement No. 101003591 NEX-GEN SIMS, and partially by the Sustainable Underground Mining (SUM) Academy Programme, under project SP14.}
\thanks{$^1$\url{https://www.youtube.com/watch?v=f9BnK34XkuQ}}
}%

\begin{document}

\maketitle
\thispagestyle{empty}
\pagestyle{empty}

\begin{abstract}
In the field of resource-constrained robots and the need for effective place recognition in multi-robotic systems, this article introduces RecNet, a novel approach that concurrently addresses both challenges. The core of RecNet's methodology involves a transformative process: it projects 3D point clouds into range images, compresses them using an encoder-decoder framework, and subsequently reconstructs the range image, restoring the original point cloud. Additionally, RecNet utilizes the latent vector extracted from this process for efficient place recognition tasks. This approach not only achieves comparable place recognition results but also maintains a compact representation, suitable for sharing among robots to reconstruct their collective maps. The evaluation of RecNet encompasses an array of metrics, including place recognition performance, the structural similarity of the reconstructed point clouds, and the bandwidth transmission advantages, derived from sharing only the latent vectors. Our proposed approach is assessed using both a publicly available dataset and field experiments$^1$, confirming its efficacy and potential for real-world applications.  
\end{abstract}


\section{INTRODUCTION} \label{sec:introduction}
%
Place recognition plays a fundamental role in both single and multi-robot Simultaneous Localization and Mapping (SLAM) systems, as it allows robots to identify previously visited locations, while mitigating the effects of drift. In these systems, the ability of robots to recognize previously visited locations, allows them to anchor their position estimates and correct accumulated errors over time~\cite{Denniston2022LoopCP}. Image-based place recognition has been widely explored~\cite{uy2018pointnetvlad}, but it suffers from certain limitations, such as sensitivity to lighting conditions and occlusions, which can restrict its reliability in challenging real-world environments. In contrast, LiDAR-based place recognition has gained popularity thanks to its ability to provide robust and precise 3D representations of the environment.
Despite the advantages of LiDAR-based methods, they come with their own set of challenges. Direct computations in the 3D point cloud data can be computationally intensive, limiting real-time performance on resource-constrained robotic platforms. Additionally, in multi-robot systems, a major challenge lies in efficiently sharing individual maps and point cloud representations among agents. Centralized communication approaches suffer from limitations in scalability and are prone to single points of failure, hindering the potential benefits of collaborative multi-robot mapping and map-merging~\cite{stathoulopoulos2023frame}. Consequently, the need arises for a more decentralized and efficient communication mechanism that allows agents to share their point cloud representations directly and selectively. With multiple robots operating over extended periods of time, the accumulation of vast amounts of point cloud data becomes inevitable~\cite{Agha2021}. This accumulation not only strains the onboard computational resources of the robots but also leads to significant communication overhead, when sharing maps among agents for collaborative mapping~\cite{Gielis2022}. Additionally, as the number of robots in a multi-robot system grows, the complexity of map-merging and coordination escalates exponentially, demanding more sophisticated and scalable solutions.
\begin{figure}[t!]
    \centering
    \includegraphics[width=0.95\columnwidth]{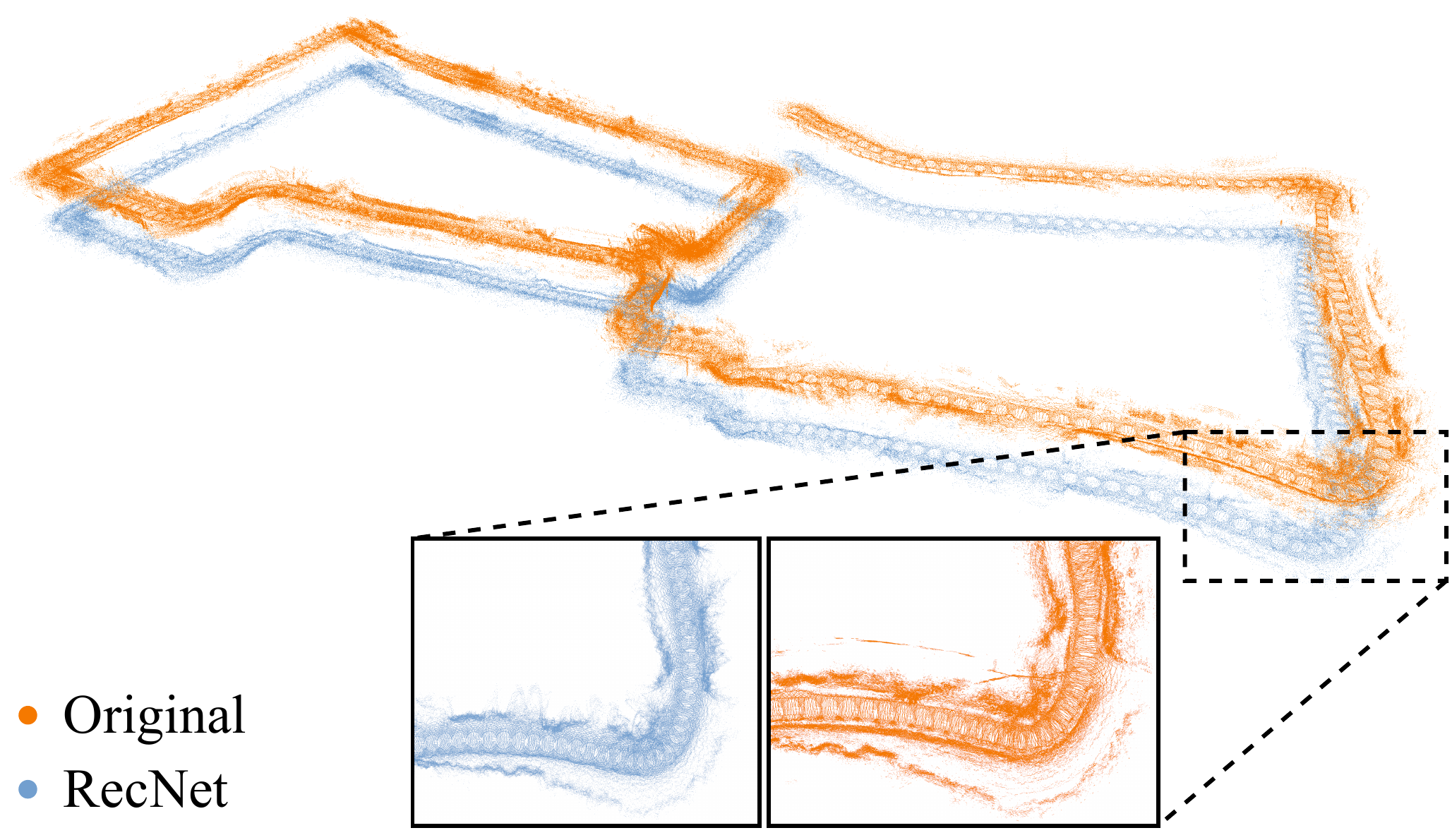}
    \setlength{\belowcaptionskip}{-18pt}
    \caption{The original and the reconstructed map using RecNet on an experiment around Lule\r{a} University of Technology.} 
    \label{fig:concept_ltu}
\end{figure}

In response to the challenges posed by resource-constrained robots and the need for effective place recognition in multi-robot systems, this article presents a novel approach that tackles both aspects simultaneously. Our method integrates LiDAR-based place recognition with lightweight descriptors, enabling efficient point cloud representation and sharing with other agents. By transforming 3D point clouds into compact range images and then lightweight descriptors, we significantly reduce computational burden and communication overhead. The central focus of this approach is to facilitate seamless data sharing~\cite{damigos2023com_aware} and to promote multi-robot collaboration in exploration missions~\cite{chang2023dlite}. This not only enhances the capabilities of individual robots but also empowers the entire multi-robot system to collectively build and maintain a comprehensive map. The efficiency of our method is crucial in resource-constrained scenarios, where computational and communication limitations can hinder the effectiveness of traditional approaches~\cite{Lindqvist2023}.

The contributions of the proposed framework can be summarized as follows: (a) Introduction of a novel descriptor pipeline that not only matches the performance of state-of-the-art approaches like those discussed in~\cite{chen2020overlapnet} and~\cite{schaupp2019oreos}, but also distinguishes itself by offering a reverse conversion to the original point cloud representation. This unique feature sets it apart from conventional methods; (b) Unlike the majority of existing approaches, RecNet provides point cloud reconstruction, while simultaneously achieves faster descriptor extraction when compared to~\cite{dube2020segmap}, a feature that opens up possibilities for real-time global localization. Furthermore, the reconstruction quality was evaluated using publicly available structural similarity metrics~\cite{AlexiouPointSSIM}, on a well-established dataset. 
(c) The descriptors generated by our framework provide a concise representation, reducing data transmission load by over 10 times. Additionally, the reconstruction process achieves a geometrical similarity of approximately 70\%. 
The proposed framework was evaluated in both the KITTI dataset~\cite{Geiger2012CVPR} and a real-world outdoor experiment.
These evaluations showcase the robustness and versatility of our approach across various environments and scenarios, offering compelling evidence of its effectiveness in addressing real-world challenges in point cloud recognition.
\section{RELATED WORK}

In this section, we first present point cloud compression methods, followed by a discussion on place recognition techniques, with an emphasis on learning-based approaches for 3D point clouds. A comprehensive review of related literature on global localization in 3D point clouds is available in~\cite{yin2023survey}.

\textbf{Point cloud compression:}
Several approaches have been proposed for compressing LiDAR point cloud data. The RIDDLE compression pipeline~\cite{Theis2022riddle}, employs quantization followed by deep delta encoding and entropy encoding to achieve efficient compression of range image data.
Feng et al.~\cite{Feng2020} present a system that exploits spatial and temporal redundancies in LiDAR point cloud sequences. By identifying key frames and utilizing overlaps between consecutive frames, their approach achieves compression ratios ranging from x40-90 while maintaining high accuracy and real-time processing.
Sun et al.~\cite{Sun2019} propose a compression algorithm that segments 3D range data into ground and main objects, leveraging shape-based prediction methods inspired by 3D video coding techniques. By efficiently compressing prediction residuals, their method achieves compression ratios of nearly 5\% without distance distortion.
Wang et al. \cite{Wang2022} propose a three-stage framework focusing on utilizing entropy modeling and a novel attention convolutional layer for feature extraction in scanning LiDAR point clouds. Their framework, employing RICNet$_{stage1}$ and RICNet$_{stage2}$ for compression and decompression, demonstrates promising results in terms of compression ratio and reconstruction quality, especially in diverse LiDAR line scenarios.

\textbf{Place recognition:} 
The most common step in 3D point cloud place recognition is to extract discriminative features. PointNetVLAD~\cite{uy2018pointnetvlad} introduced an end-to-end trainable global descriptor by extracting local features leveraging PointNet~\cite{charles2017pointnet} and then using the NetVLAD aggregator~\cite{arandjelovic2016vlad}, leading to global descriptors for LiDAR-based place recognition. However, the limited descriptive power of PointNet prompted the proposal of LPD-Net~\cite{liu2019lpdnet} as an alternative approach. More recently, MinkLoc3D~\cite{komorowski2021mink} demonstrated superior efficiency and performance compared to previous methods by employing sparse convolutions to effectively capture point-level features. On the other hand, LoGG3D-Net~\cite{vidanapathirana2022logg3d} improved global descriptors with a local consistency loss, ensuring feature consistency within point clouds. SegMatch~\cite{dube2017segmatch} and SegMap~\cite{dube2020segmap} introduced a high-level perception technique that segments point clouds into distinct and discriminative elements at the object level. To achieve this, they utilize a 3D CNN to encode segment features and identify candidate correspondences through k-nearest neighbours (kNN) in feature space. This approach provides both the ability for place recognition and reconstructing the segments back to point clouds, inspiring the development of our approach. However, segmentation approaches may suffer from limitations in various environments, where complex or dynamic scenes can lead to inaccuracies in segment boundaries, affecting recognition performance. 

Recognizing the limitations of both segment-based and direct point cloud computations, the research community emerged with alternative solutions that focus on converting the 3D point cloud representation into the image space. The work in~\cite{elbaz2017autoencoder} introduced the LORAX algorithm for point cloud registration in localization, involving division into super-points with Random Sphere Cover Set, projection onto 2D depth maps, dimension reduction using a Deep Neural Network Auto-Encoder, and coarse registration with iterative fine-tuning. Nevertheless, the authors acknowledged that their implementation is not optimized for real-time operation, and no timings were provided. To improve computational efficiency, many methods dealing with larger scans opt to convert them into an intermediary representation prior to processing with a deep neural network. 
The work in~\cite{schaupp2019oreos} presented OREOS, designed for place recognition, which additionally estimates the yaw discrepancy between scans. Their approach involves projecting the input data into a 2D range image using a spherical projection model. Taking the concept of using range images further, OverlapNet~\cite{chen2020overlapnet} takes advantage of multiple cues such as range, normals, intensity, and semantic classes. These cues are projected as spherical images, extracted from the point cloud, thereby enhancing the framework's performance. A more advanced version, OverlapTransformer~\cite{ma2022overlaptrans}, offers a rotation-invariant representation and faster inference capabilities, although it lacks the ability to estimate yaw angles. Another notable approach, LiDAR Iris~\cite{wang2020iris}, encodes the height information of a 3D point cloud into a binary image, transforming it into the Fourier domain to achieve rotation invariance.

\section{THE PROPOSED APPROACH}

The scope of this article is to develop a novel pipeline for compressing 3D point cloud information with the purpose of: (a) performing place recognition related tasks, like loop closure~\cite{Denniston2022LoopCP} or map-merging~\cite{stathoulopoulos2023frame}, and (b) being able to reconstruct the 3D point cloud to its original form. A critical constraint imposed on this pipeline is the requirement for the compressed data to have a substantially reduced size compared to the original, ensuring efficient transmission between robots, while minimizing communication network congestion. Therefore, given a point cloud $\mathcal{P} \in \mathbb{R}^3$, we are looking for a series of functions that will compress the information into a compact representation $\beta \in \mathbb{R}^N$. The vector $\beta$ needs to be able to provide enough information in order to recover most of the spatial information from $\mathcal{P}$ into the new reconstructed point cloud $\hat{\mathcal{P}} \in \mathbb{R}^3$. In this article, we assume that the fourth dimension of the original point cloud, the intensity, is not recovered, as we keep the focus on the spatial information. The proposed approach is split into multiple modules, as described in the following subsections.
\begin{figure*}[t!]
    \centering
    \includegraphics[width=\textwidth]{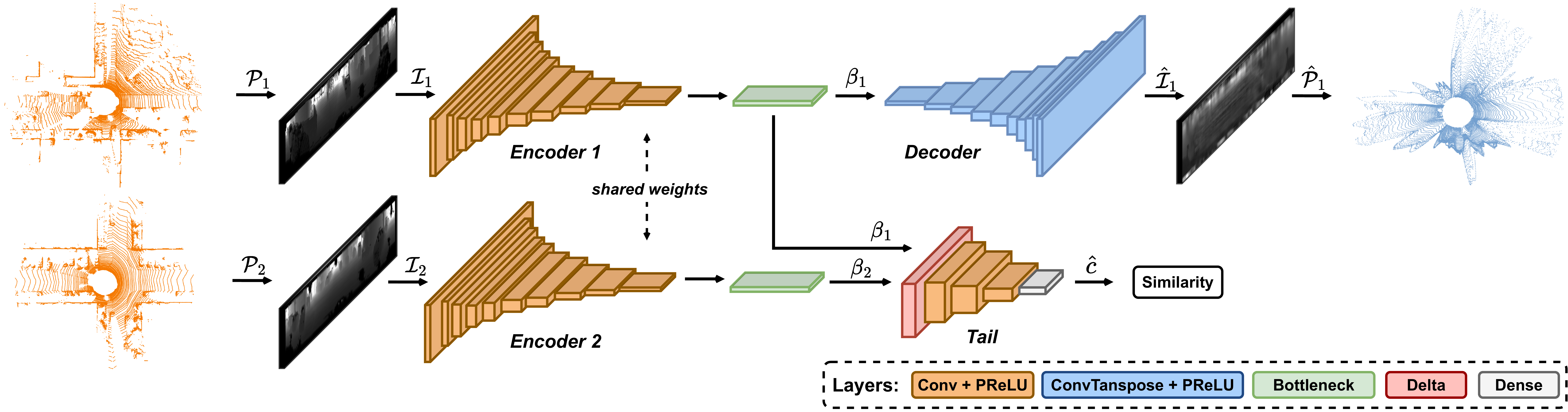}
    \setlength{\abovecaptionskip}{-4pt}
    \setlength{\belowcaptionskip}{-14pt}
    \caption{The overall RecNet pipeline is composed of two identical \textit{encoder} legs that share weights, a single \textit{decoder} leg and a \textit{tail} network, responsible for estimating the similarity between the two latent bottleneck vectors.}
    \label{fig:architecture}
\end{figure*}
\vspace{-0.4cm}
\subsection{3D Point cloud to Range Image} \label{subsec:pcd2ri}

We utilize spherical projections of LiDAR scans as our input data. These projections are commonly employed to accelerate computations~\cite{chen2020overlapnet, ma2022overlaptrans}. Our approach involves transforming the point cloud $\mathcal{P}$ into what is usually called a \textit{vertex} map $\mathcal{V}: \mathbb{R}^2 \rightarrow \mathbb{R}^3$. In this map, each pixel corresponds to the nearest 3D point. To achieve this, every point $p_i = (x, y, z) \in \mathbb{R}^3$ undergoes a conversion process. Initially, it is transformed using the function $\Pi: \mathbb{R}^3 \rightarrow \mathbb{R}^2$ into spherical coordinates and subsequently mapped to image coordinates $I=(u, v) \in \mathbb{R}^2$. This process can be summarized as follows:
\begin{equation} \label{eq:pcd2ri}
    \Bigg(\begin{array}{c}
       u  \\
       v  
    \end{array}\Bigg) = 
    \Bigg(\begin{array}{c}
        \frac{1}{2}\left[1 - \arctan{(yx^{-1})} \pi^{-1}\right]w \\
        \left[1 - (\arcsin{(zr^{-1})} + \text{f}_{\text{up}}) \text{f}^{-1}\right]h
    \end{array} \Bigg),
\end{equation}
where $r = ||p||_2$ is the range, $\text{f} = \text{f}_{\text{up}} + \text{f}_{\text{down}}$ corresponds to the vertical field-of-view and $w$, $h$ denote the width and height dimensions of the resultant vertex map $\mathcal{V}$.

\subsection{The Encoder-Decoder pair} \label{subsec:encoder-decoder}

Following the initial compression of three-dimensional data onto a two-dimensional plane, we apply a subsequent compression step utilizing an \textit{encoder}. This \textit{encoder} takes a range image $I$ as input and generates a multidimensional descriptive vector $\mathcal{E}(I) = \beta \in \mathbb{R}^{256\times64}$. Immediately thereafter, the bottleneck $\beta$ is passed to a \textit{decoder} with the objective of reconstructing the image $I$. The \textit{decoder} can be conceptualized as the approximate inverse function of the \textit{encoder}, aiming to produce a reconstructed image, $\hat{I} \in \mathbb{R}^2$. This can be formally defined as $\mathcal{D}(\beta) = \inv{\tilde{\mathcal{E}}}(\beta) = \hat{I}$. The role of $\beta$ is to encapsulate sufficient descriptive information to facilitate image restoration.
As a final step, in line with our earlier objective (a) (performing place recognition tasks), we identify potential candidates, by assessing the similarity between each query vector $\beta_1$, and the corresponding map vectors $\beta_2$. We achieve this by passing the bottlenecks through a smaller Convolutional Neural Network (CNN) to get an estimate on the similarity of the two latent vectors, defined as $\mathcal{T}(\beta_1, \beta_2) = \hat{c} \in [0,1]$.
For a more comprehensive understanding of the network architecture and the loss functions, we provide detailed information in sub-section~\ref{subsec:network_architecture}.

\subsection{Range Image to 3D Point cloud} \label{subsec:ri2pcd}

The final stage of the process involves the conversion of the predicted image $\hat{I}$ back into a three-dimensional representation, denoted as $\hat{\mathcal{P}} \in \mathbb{R}^3$. This conversion can be understood as the approximate inverse mapping of the transformation, detailed in Section~\ref{subsec:pcd2ri}, represented by the mapping function $\inv{\tilde{\Pi}}: \mathbb{R}^2 \rightarrow \mathbb{R}^3$. Solving Eq. (\ref{eq:pcd2ri}) for $x$, $y$, and $z$ given $r$ from the pixel intensity, is straightforward. It is worth noting that the original transformation $\Pi$ operates from a three-dimensional space to a two-dimensional one, resulting in a loss of information due to the quantization of the image grid and is not a fully inversable and~1-1 mapping. Consequently, it is expected that the resulting point cloud will serve as an approximation of the original point cloud due to this inherent loss of information.


\subsection{The Network Architecture} \label{subsec:network_architecture}

The proposed network architecture is illustrated in Fig.~\ref{fig:architecture}. To optimize the network for lightweight and rapid processing, we exclusively leverage the depth information from the point cloud to generate input range images, which also serve as the target images for our reconstruction task. This choice is motivated by our desire to maintain the network's efficiency, as employing normals images would imply additional computation time for computing neighboring relations. Furthermore, in line with findings from~\cite{chen2020overlapnet}, it has been observed that the incorporation of intensity, normal, and semantic images has a negligible impact on the network's performance, except for the yaw estimation aspect, which falls beyond the scope of this article. As a result, the input for the \textit{encoder} is a tensor with dimensions $1\times64\times900$, 
with the channel, height, and width, respectively. In this tensor, the single channel represents the range information, as previously mentioned.
RecNet adopts a Siamese network design, comprising two \textit{encoder} legs, a single \textit{tail}, and a single \textit{decoder} leg. Notably, the \textit{encoders} are constructed to be identical, sharing weights across their respective counterparts. Further insights into the trainable parameters are available in Table~\ref{table:network_architecture}.

\textbf{Encoder:} The \textit{encoder} is composed of two legs, all possessing an identical architecture and weight sharing. Each leg represents a fully convolutional network (FCN) featuring ten convolutional layers. We abstain from incorporating max-pooling layers, as they tend to erode vital positional information, a critical element in the image reconstruction process. This architecture maintains a degree of lightness, generating feature volumes sized at $256\times1\times64$, denoted as $\beta$. The inherent translation-equivariance of the FCN permits the future application of yaw discrepancy estimation through a correlation head~\cite{chen2020overlapnet}.

\textbf{Decoder:} The \textit{decoder}, featuring a single leg, mirrors the architecture of the \textit{encoder}. However, 2D convolutions are replaced by ten transpose 2D convolutions, facilitating the restoration of the original height and width of the image, while reducing the number of channels. This module operates akin to the gradient of a 2D convolution concerning its input. While commonly referred to as a fractionally-strided convolution or deconvolution, it does not perform a true inverse of convolution. The \textit{decoder}'s output directly yields the reconstructed range image in the original size.

\textbf{Tail:} The \textit{tail} is a straightforward CNN comprising three convolutional layers and a fully connected layer. It takes two bottleneck vectors as input, and prior to entering the convolutional layers, these vectors undergo preprocessing through a \textit{delta} layer that involves subtracting one from the other, resulting in a new vector denoted as $\Delta(\beta_1, \beta_2) = \beta_1 - \beta_2 \in \mathbb{R}^{256\times64}$. Subsequently, the convolutional layers follow to extract prominent features that aid in distinguishing between the latent vectors. Lastly, the fully connected layer quantifies the similarity between the two vectors, offering a confidence level for their similarity.

\begin{table}
\centering
\caption{The layers of the RecNet architecture. Every operator is followed by a PReLU activation function, except the last layers of the \textit{encoder} and \textit{decoder}. A batch normalization is added every other \texttt{Conv2D} and \texttt{ConvTran2D}.}
\label{table:network_architecture}
\begin{tblr}{
  width = \linewidth,
  colspec = {Q[27]Q[184]Q[142]Q[121]Q[154]Q[294]},
  row{odd} = {c},
  row{2} = {c},
  row{4} = {c},
  row{6} = {c},
  row{8} = {c},
  row{10} = {c},
  row{12} = {c},
  row{14} = {c},
  row{16} = {c},
  row{18} = {c},
  row{20} = {c},
  row{24} = {c},
  row{26} = {c},
  cell{2}{1} = {r=10}{},
  cell{12}{1} = {r=10}{},
  cell{22}{1} = {r=4}{},
  cell{22}{2} = {c},
  cell{22}{3} = {c},
  cell{22}{4} = {c},
  cell{22}{5} = {c},
  cell{22}{6} = {c},
  hline{1,26} = {-}{0.08em},
  hline{2,12,22} = {-}{0.05em},
}
                                      & Operator~       & Stride  & Features & Kernel   & Output Shape          \\
\begin{sideways}Encoder\end{sideways} & Conv2D          & $(2,2)$ & $16$     & $(5,15)$ & $16\times30\times443$ \\
                                      & Conv2D          & $(2,2)$ & $16$     & $(3,15)$ & $16\times14\times215$ \\
                                      & Conv2D          & $(2,2)$ & $32$     & $(3,13)$ & $32\times6\times102$  \\
                                      & Conv2D          & $(2,1)$ & $32$     & $(2,12)$ & $32\times3\times90$   \\
                                      & Conv2D          & $(2,1)$ & $64$     & $(2,9)$  & $64\times1\times82$   \\
                                      & Conv2D          & $(1,1)$ & $64$     & $(1,7)$  & $64\times1\times76$   \\
                                      & Conv2D          & $(1,1)$ & $128$    & $(1,5)$  & $128\times1\times72$  \\
                                      & Conv2D          & $(1,1)$ & $128$    & $(1,5)$  & $128\times1\times68$  \\
                                      & Conv2D          & $(1,1)$ & $256$    & $(1,3)$  & $256\times1\times66$  \\
                                      & Conv2D          & $(1,1)$ & $256$    & $(1,3)$  & $256\times1\times64$  \\
\begin{sideways}Decoder\end{sideways} & ConvTran2D & $(1,1)$ & $256$    & $(1,3)$  & $256\times1\times66$  \\
                                      & ConvTran2D & $(1,1)$ & $128$    & $(1,3)$  & $128\times1\times68$  \\
                                      & ConvTran2D & $(1,1)$ & $128$    & $(1,5)$  & $128\times1\times70$  \\
                                      & ConvTran2D & $(1,1)$ & $64$     & $(1,5)$  & $64\times1\times74$   \\
                                      & ConvTran2D & $(1,1)$ & $64$     & $(1,7)$  & $64\times1\times82$   \\
                                      & ConvTran2D & $(2,1)$ & $32$     & $(2,9)$  & $32\times2\times90$   \\
                                      & ConvTran2D & $(2,1)$ & $32$     & $(2,12)$ & $32\times5\times102$  \\
                                      & ConvTran2D & $(2,2)$ & $16$     & $(3,13)$ & $16\times13\times215$ \\
                                      & ConvTran2D & $(2,2)$ & $16$     & $(3,15)$ & $16\times30\times443$ \\
                                      & ConvTran2D & $(2,2)$ & $1$      & $(5,15)$ & $1\times64\times900$  \\
\begin{sideways}Tail\end{sideways}    & Conv2D          & $(5,5)$ & $32$    & $(9,9)$  & $32\times50\times50$  \\
                                      & Conv2D          & $(3,3)$ & $64$     & $(5,5)$  & $64\times16\times16$   \\
                                      & Conv2D          & $(1,1)$ & $128$     & $(3,3)$  & $128\times14\times14$   \\
                                      & Dense           & -       & -        & -        & $1$     
\end{tblr}
\end{table}
\begin{figure}[t!]
    \centering
    \includegraphics[width=1.0\columnwidth]{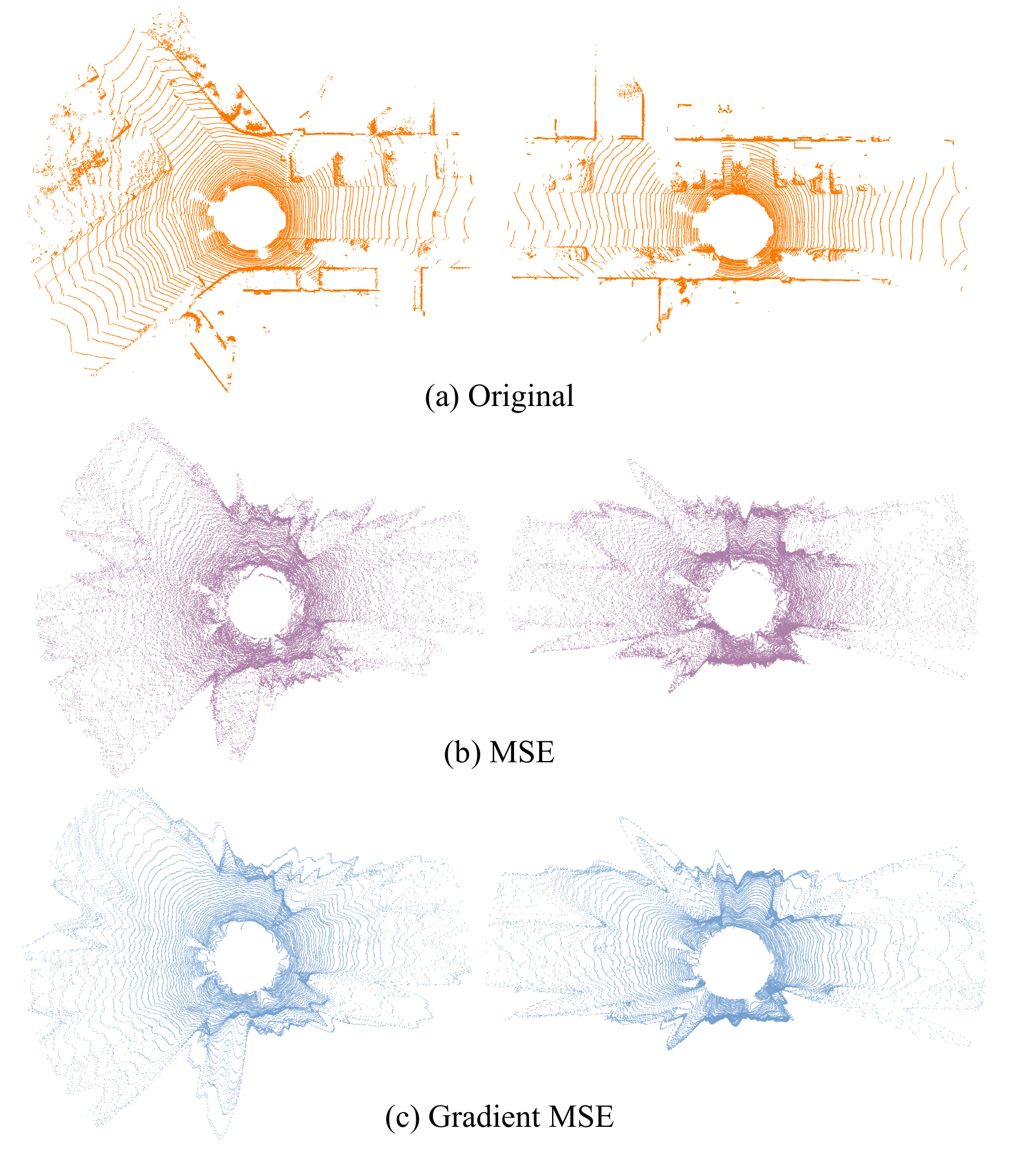}
    \setlength{\abovecaptionskip}{-12pt}
    \setlength{\belowcaptionskip}{-17pt}
    \caption{Two LiDAR scans from the KITTI dataset, where (a) depicts the original scan, (b) is the reconstructed using the MSE and (c) is the reconstructed using the Gradient MSE.}
    \label{fig:scan_comparison}
\end{figure}

\subsection{Loss functions} \label{subsec:loss_functions}
We adopt an end-to-end training approach to achieve two key objectives: reconstructing the original range image and learning compact representations for tasks related to place recognition. These objectives are governed by two distinct loss terms: the reconstruction loss ($L_{rec}$) and the place recognition loss ($L_{pr}$), with a weighting factor $\alpha$ used to combine them effectively. This combination is expressed as:
\begin{equation}
    L(I, \hat{I}, \beta_1, \beta_2) = L_{rec}(I,\hat{I}) + \alpha L_{pr}(\beta_1, \beta_2)
\end{equation} 
To begin with, the reconstruction loss ($L_{rec}$) employs a custom loss function called Weighted Pixel Gradient MSE. This function blends the conventional pixel-wise Mean Squared Error (MSE) loss ($L_{MSE}$) with a weighted loss component accounting for the gradient differences between corresponding pixels in both images ($L_{\partial}$). The MSE is defined as:
\begin{equation}
    L_{MSE}(I,\hat{I}) = \frac{1}{N} \sum^N_{i=0} (I_i-\hat{I_i})^2
\end{equation}
The Gradient MSE loss is a component of the custom loss function designed to measure the difference between the gradients of the reconstructed image and the gradients of the target image. It serves as a way to encourage the model to generate images with similar local edge and texture features as the target image. The Gradient MSE loss is defined as:
\begin{equation}
    L_{\partial}(I,\hat{I}) = \frac{1}{N} \bigg( \sum_{i=0}^N |\frac{\partial I_i}{\partial u} - \frac{\partial \hat{I_i}}{\partial u}| + \sum_{i=0}^N |\frac{\partial I_i}{\partial v} - \frac{\partial \hat{I_i}}{\partial v}| \bigg)
\end{equation}
Initially, the gradients of both the reconstructed image and the target image are determined. These gradients reflect how pixel values change in both horizontal and vertical directions. Subsequently, the absolute disparities between the gradients of the reconstructed image and the target image are calculated. These calculations are performed separately for the $u$-direction and the $v$-direction. The absolute gradient differences are summed across all pixels in the image, and finally the total gradient difference is divided by the total number of pixels ($N$) to compute the average absolute gradient difference.
The gradient difference loss encourages the model to pay attention to fine details and textures in the image, as it penalizes discrepancies in how pixel values change from one location to another compared to the target image. This is particularly useful in image generation tasks where maintaining fine-grained structures and edges is important, such as denoising, super-resolution, and image reconstruction.
Moving to the place recognition loss, we have devised a custom loss function with the primary objective of training the model to predict a similarity score between the bottleneck vectors corresponding to the query point clouds and those of the map. This similarity score is calculated as the Euclidean distance between the two corresponding poses but it is transformed using an exponential function to constrain its range between 0 and 1. We approach this as a simple regression problem, where our loss function aims to minimize the mean square error between the network's predicted similarity score $\hat{c}$, and the target similarity score, defined as follows:
\begin{equation}
    L_{pr}(p_1, p_2, \hat{c}) = \big(C(p_1, p_2) - \hat{c}\big)^2,
\end{equation}
where, $C(p_1, p_2) = \big(\exp(-d(p_1,p_2)/m)\big)^{-1}$, represents the target similarity score, and $d(\cdot)$ calculates the Euclidean distance between the two poses, $p_1$ and $p_2$. The parameter $m$ serves as a threshold that defines the desired scale for the similarity measurement.

\section{EXPERIMENTAL EVALUATION} \label{sec:experimental_eval}
\begin{figure}[b]
    \centering
    \includegraphics[width=1\columnwidth]{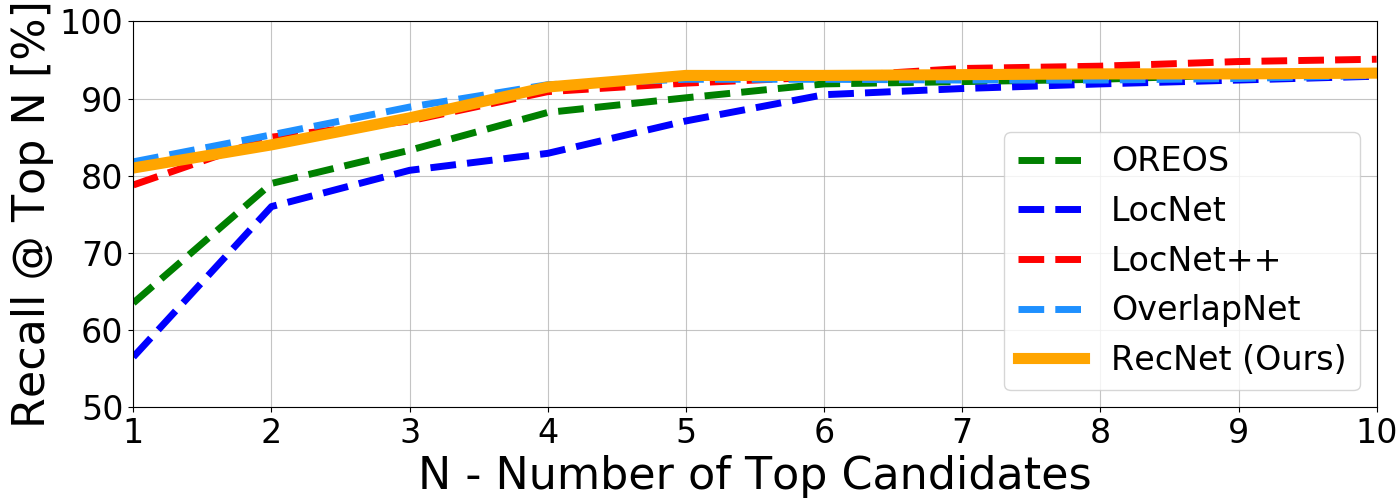}
    \setlength{\abovecaptionskip}{-5pt}
    \caption{The place recognition performance of RecNet compared to others, in the 00 sequence of KITTI.}
    \label{fig:pr_results}
\end{figure}
The experimental evaluation validates our approach, focusing on two main aspects: place recognition and point cloud reconstruction. Initially, we assess our method using KITTI data~\cite{Geiger2012CVPR}, which includes LiDAR scans of urban areas in Germany. Our setup follows the previous works~\cite{chen2020overlapnet, schaupp2019oreos}, with sequences 03-10 for training, sequence 02 for validation, and sequence 00 for evaluation.
To test our method's generalization, we collected LiDAR scans using an Ouster OS1-32 while driving around Luleå University of Technology, referred to as LTU. This scanner has fewer beams, resulting in smaller projected images ($1\times32\times450$). To adapt, we adjusted the network, setting the bottleneck size to $\beta_{mini} \in \mathbb{R}^{128\times32}$ and removing the last two convolutional layers of the encoder and decoder.
RecNet is implemented in \texttt{PyTorch}, and training was conducted on a workstation with an NVIDIA GeForce RTX 4090 graphics card and a 13th Gen Intel Core i9-13900K. All experiments were performed on a laptop with an 11th Gen Intel Core i7-1165G7, even without GPU support. The preprocessing time needed for a single point cloud scan is approximately 78ms, while the encoding for the bottleneck vector takes just 2ms. 
\begin{figure*}[!t]
    \centering
    \includegraphics[width=\textwidth]{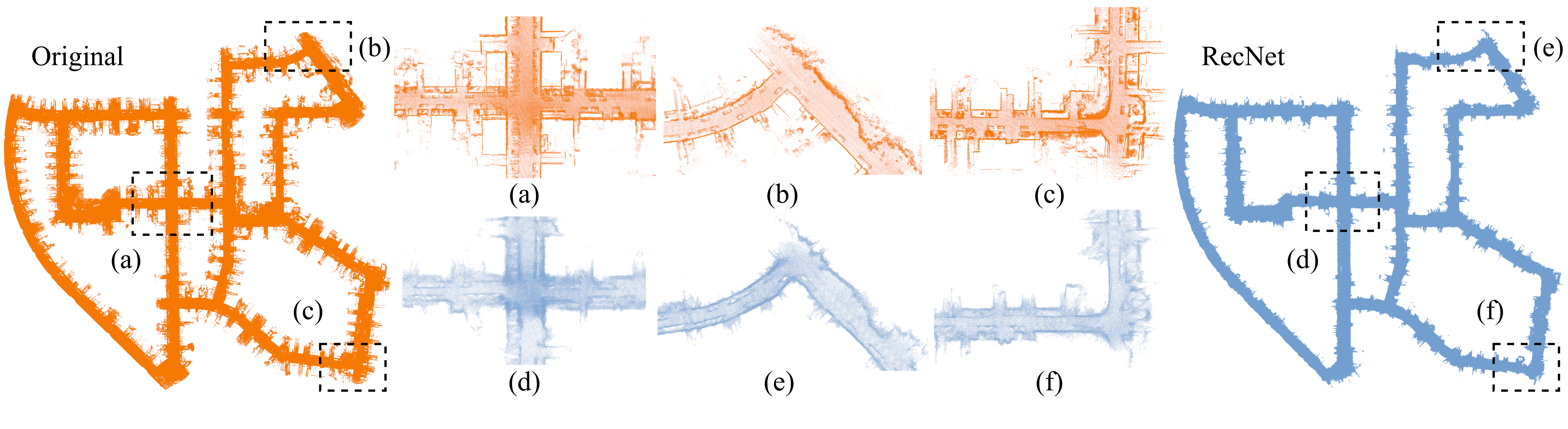}
    \setlength{\abovecaptionskip}{-12pt}
    \setlength{\belowcaptionskip}{-10pt}
    \caption{The original and reconstructed point cloud maps from the 00 sequence of KITTI. The highlighted segments (a) - (c) provide a closer look at the original map, while the segments (d) - (f) demonstrate the corresponding reconstructed parts. RecNet facilitates a reconstruction that resembles the original map by only transmitting the latent vectors of the network.}
    \label{fig:map_segments}
\end{figure*}

\subsection{Place Recognition Performance} \label{subsec:pr_results}
This section evaluates our approach's effectiveness in place recognition and loop closure candidate identification. We utilize a 10-meter threshold for the loss function, resulting in a similarity score of about 0.6 for scans at this distance. For candidate classification, we employ a 0.75 threshold, that corresponds to approximately a 3-meter radius.
To assess candidate retrieval, we use sequence 00 from the KITTI dataset. We designate the initial 170 seconds, along with the corresponding predicted bottleneck vectors, as map vectors. The remaining data is employed for querying potential candidates. The obtained results, depicted in Fig.~\ref{fig:pr_results}, indicate performance comparable to or surpassing state-of-the-art lightweight approaches such as LocNet~\cite{GidarisK16}, OREOS~\cite{schaupp2019oreos} and OverlapNet~\cite{chen2020overlapnet}.
Notably, our bottleneck vector $\beta$ is not only smaller than OverlapNet's (approximately 5.5 times smaller) but also retains the ability to reconstruct the point cloud scan. Better performance than OREOS is expected due to the similarity in the encoding process, however RecNet has a greater number of convolutional layers and a larger latent vector, yielding a deeper feature extraction process.
\begin{table}[b]
\centering
\caption{Comparison of the PointSSIM~\cite{AlexiouPointSSIM} similarity score, for different reconstruction loss functions.}
\label{table:similarity}
\begin{tblr}{
  width = \linewidth,
  colspec = {Q[5]Q[220]Q[170]Q[170]Q[170]Q[170]},
  row{1} = {c},
  row{3} = {c},
  row{4} = {c},
  row{6} = {c},
  row{7} = {c},
  cell{2}{1} = {r=3}{},
  cell{2}{2} = {c},
  cell{2}{3} = {c},
  cell{2}{4} = {c},
  cell{2}{5} = {c},
  cell{2}{6} = {c},
  cell{5}{1} = {r=3}{},
  cell{5}{2} = {c},
  cell{5}{3} = {c},
  cell{5}{4} = {c},
  cell{5}{5} = {c},
  cell{5}{6} = {c},
  hline{1,8} = {-}{0.08em},
  hline{2,5} = {-}{0.05em},
}
                                    & Method                              & Corr@$0.5$m [$\%$]             & GeomSim [$\%$]                 & NormSim [$\%$]                 & CurvSim [$\%$]                 \\
\begin{sideways}KITTI\end{sideways} & Downsampled                                 & $\textbf{76.2}\pm\textbf{2.1}$                    & $23.4\pm1.6$                   & $80.3\pm1.3$                   & $17.7\pm2.8$                                \\
                                    & $\mathcal{L}_{MSE}$                        & $68.8\pm2.7$                   & $64.2\pm2.5$                   & $80.7\pm1.1$                   & $25.0\pm2.5$                   \\
                                    & $\mathcal{L}_{MSE} + \mathcal{L}_\partial$ & $69.2\pm2.4$ & $\textbf{65.9}\pm\textbf{2.5}$ & $\textbf{81.4}\pm\textbf{1.3}$ & $\textbf{27.8}\pm\textbf{1.9}$ \\
\begin{sideways}LTU\end{sideways}   & Downsampled                                 & $\textbf{75.6}\pm\textbf{2.7}$                    & $25.9\pm1.7$                   & $65.9\pm2.4$                    & $19.5\pm2.6$                              \\
                                    & $\mathcal{L}_{MSE}$                        & $65.8\pm3.0$                   & $68.7\pm2.6$                   & $67.0\pm2.2$                   & $36.4\pm2.8$                   \\
                                    & $\mathcal{L}_{MSE} + \mathcal{L}_\partial$ & $67.3\pm2.9$ & $\textbf{71.1}\pm\textbf{2.6}$ & $\textbf{67.8}\pm\textbf{1.9}$ & $\textbf{38.9}\pm\textbf{2.4}$ 
\end{tblr}
\end{table}
\subsection{Point cloud Reconstruction} \label{subsec:reconstruction_results}

The second phase of the evaluation examines how well RecNet reconstructs the original point cloud, gauging its fidelity and precision in preserving the essential details. In order to evaluate the reconstruction, we use four metrics. The first is a simple correspondence percentage at 0.5 meters. This metric is not very descriptive of the ability of the scan to maintain certain aspects like geometrical shape, curvatures and so on. For that we present the three metric scores proposed in~\cite{AlexiouPointSSIM}, which  measure the geometric, curvature and normal vectors' similarity. In addition to the reconstructed scans, we include to the comparison a sub-sampled version of the scans. We downsample with a voxel size of 0.75, in order to match the bottleneck vectors' flattened size . The scores presented in Table~\ref{table:similarity} correspond to the mean and standard deviation of the average of all the scans in the sequence 00 of KITTI and all the LiDAR scans attained from the LTU drive. 
In Fig.~\ref{fig:scan_comparison}, we present the original and reconstructed point clouds using our proposed pixel-wise Gradient MSE and the simple MSE. Clearly, the gradient approach preserves line clarity and closely resembles the original point cloud.
Fig.~\ref{fig:map_segments} and~\ref{fig:concept_ltu} illustrate the results of registering individual point cloud scans based on odometry, for the KITTI dataset and the LTU drive, respectively. Our reconstructed map retains the overall shape and information of the original map. This promising outcome paves the way for evaluating its applicability in real-world multi-robot missions, showcasing its potential for practical use.
\vspace{-5pt}
\subsection{Point cloud Transmission} \label{subsec:pcd_transmission}
A key objective is to achieve an efficient representation for inter-robot communication, therefore the third aspect measures the reduction in point cloud transmission bandwidth afforded by our approach. This metric serves as a tactile measure of the real-world viability and applicability of our methodology.
In Table~\ref{table:bandwidth}, we present a comparative analysis of the bandwidth requirements for transmitting downsampled original point clouds (where the downsampling matches established SLAM practices~\cite{liosam2020shan}) against the bandwidth necessary for transmitting only the bottleneck vectors. The results demonstrate the substantial reduction in bandwidth demands achieved through RecNet, with reductions exceeding tenfold in both the KITTI and the LTU drive experiments. This reduction emphasizes the efficiency and potential resource-saving benefits that our method brings to the front in the context of multi-robot systems.
\begin{table}[!t]
\centering
\caption{Statistics for the transmission of the point clouds and the corresponding bottleneck vectors.}
\label{table:bandwidth}
\begin{tblr}{
  width = \linewidth,
  colspec = {Q[887]Q[95]Q[131]},
  column{2} = {c},
  column{3} = {c},
  hline{1-2,7} = {-}{0.08em},
}
\textbf{Statistics:}                     & \textbf{KITTI} & \textbf{LTU} \\
Mission Duration (s)                     & 470            & 368          \\
Bandwidth for transmitting original clouds (kB/s) & 2,560         & 1,220       \\
Bandwidth for bottleneck vectors (kB/s)  & 131            & 88           \\
Final map size from original clouds (MB)    & 897            & 204          \\
Final size of bottleneck vectors (MB)    & 62             & 33           
\end{tblr}
\vspace{-0.5cm}
\end{table}
\section{CONCLUSIONS} \label{sec:conclusions}
In conclusion, RecNet is a novel framework for extracting descriptive information from point clouds, with place recognition performance akin to state-of-the-art approaches. Moreover, RecNet has the capacity for point cloud reconstruction, yielding substantial bandwidth reductions, particularly beneficial in the context of multi-robot systems. The evaluation of structural similarity between original and reconstructed point clouds, using a state-of-the-art metric on an openly accessible dataset, reinforces its reliability. At last, RecNet paves the way for extensive exploration of its utility in autonomous multi-robot systems, spanning from navigation, localization, map-merging, and beyond.

\addtolength{\textheight}{-5.5cm}   




\bibliographystyle{./IEEEtranBST/IEEEtran}
\bibliography{./IEEEtranBST/IEEEabrv,revised}


\end{document}